%% file: iclr2026_conference.tex
\documentclass{article} 
\usepackage{iclr2026_conference,times}
\iclrfinalcopy
\input{math_commands.tex}

\usepackage{hyperref}
\usepackage{url}
\usepackage[utf8]{inputenc} 
\usepackage[T1]{fontenc}    
\usepackage{amsfonts}       
\usepackage{nicefrac}       
\usepackage{microtype}      
\usepackage{xcolor}         
\usepackage{subcaption}
\usepackage{booktabs}
\usepackage{amsmath}
\usepackage{multirow}
\usepackage{adjustbox}
\usepackage{caption}
\usepackage{wrapfig}
\usepackage{xspace} 
\newcommand{\fate}{\textsc{FATE}\xspace}
\newcommand{\mae}{\textsc{MAE}\xspace}
\newcommand{\mse}{\textsc{MSE}\xspace}

\newcommand{\cnn}{\textsc{CNN}\xspace}
\newcommand{\lstm}{\textsc{LSTM}\xspace}

\usepackage{amssymb}
\usepackage{mathtools}
\usepackage{amsthm}
\usepackage{array}       
\usepackage{placeins}
\usepackage{diagbox}
\usepackage{colortbl}

\usepackage[english]{babel}
\usepackage{blindtext}
\usepackage[utf8]{inputenc}
\usepackage[T1]{fontenc}
\usepackage{hyperref}
\usepackage{subcaption}
\usepackage{float}

\usepackage{xcolor}
\usepackage{multirow}
\usepackage{caption}
\usepackage{indentfirst}

\title{FATE: Focal-modulated Attention Encoder for Multivariate Time-series Forecasting}

\author{
Tajamul Ashraf$^{1}$, Janibul Bashir$^{1}$ \\
$^{1}$GAASH Research Lab \\
Department of Information Technology \\
National Institute of Technology Srinagar, India
}
\begin{document}

\input{preamble}

\maketitle


\input{ICLR/sec/0_abstract}
\input{ICLR/sec/1_intro}

\input{ICLR/sec/2_background_and_related_work}
\input{ICLR/sec/3_purpose_meth}
\input{ICLR/sec/4_experiments}

\input{ICLR/sec/5_conclusion}

\bibliography{iclr2026_conference}
\bibliographystyle{iclr2026_conference}


\end{document}

%% file: math_commands.tex
\usepackage{amsmath,amsfonts,bm}

\def\eqref#1{equation~\ref{#1}}

\def\1{\bm{1}}

\DeclareMathAlphabet{\mathsfit}{\encodingdefault}{\sfdefault}{m}{sl}
\SetMathAlphabet{\mathsfit}{bold}{\encodingdefault}{\sfdefault}{bx}{n}



%% file: preamble.tex
\newcommand{\ghg}{\texttt{GHG}\xspace}
\newcommand{\rnn}{\texttt{RNN}\xspace}
\newcommand{\lstnet}{\texttt{LSTNet}\xspace}
\newcommand{\gis}{\texttt{GIS}\xspace}
\newcommand{\arcgis}{\texttt{ArcGIS}\xspace}
\newcommand{\focalnet}{\texttt{Focalnet}\xspace}
\newcommand{\ann}{\texttt{ANN}\xspace}
\newcommand{\ai}{\texttt{AI}\xspace}
\newcommand{\asfod}{\texttt{A2SFOD}\xspace}
\newcommand{\irg}{\texttt{IRG}\xspace}
\newcommand{\fedavg}{\texttt{FedAvg}\xspace}
\newcommand{\msta}{\texttt{MSTA}\xspace}
\newcommand{\nlp}{\texttt{NLP}\xspace}
\newcommand{\pets}{\texttt{PETS}\xspace}
\newcommand{\sota}{\texttt{SOTA}\xspace}
\newcommand{\ckd}{\texttt{CKD}\xspace}
\newcommand{\vfms}{\texttt{VFMs}\xspace}
\newcommand{\vfm}{\texttt{VFM}\xspace}
\newcommand{\fms}{\texttt{FMs}\xspace}
\newcommand{\fm}{\texttt{FM}\xspace}
\newcommand{\cf}{\texttt{C2F}\xspace}
\newcommand{\cb}{\texttt{C2B}\xspace}
\newcommand{\kc}{\texttt{K2C}\xspace}
\newcommand{\ii}{\texttt{In2IB}\xspace}
\newcommand{\id}{\texttt{In2D}\xspace}
\newcommand{\mi}{\texttt{MI}\xspace}
\newcommand{\bcd}{\texttt{BCD}\xspace}
\newcommand{\cc}{\texttt{CC}\xspace}
\newcommand{\mlo}{\texttt{MLO}\xspace}
\newcommand{\fpi}{\texttt{FPI}\xspace}
\newcommand{\froc}{\texttt{FROC}\xspace}
\newcommand{\ccpm}{\texttt{CCPM}\xspace}
\newcommand{\ap}{\texttt{AP}\xspace}
\newcommand{\map}{\texttt{mAP}\xspace}
\newcommand{\ffdm}{\texttt{FFDM}\xspace}
\newcommand{\birads}{\texttt{BIRADS}\xspace}
\newcommand{\di}{\texttt{D2In}\xspace}
\newcommand{\dib}{\texttt{D2IB}\xspace}
\newcommand{\fused}{\texttt{FUSED}\xspace}
\newcommand{\acr}{\texttt{ACR}\xspace}
\newcommand{\DNN}{\texttt{DNN}\xspace}
\newcommand{\DNNS}{\texttt{DNNs}\xspace}
\newcommand{\SOTA}{\texttt{SOTA}\xspace}
\newcommand{\RECE}{\texttt{RECE}\xspace}
\newcommand{\RECEG}{\texttt{RECE-G}\xspace}
\newcommand{\RECEM}{\texttt{RECE-M}\xspace}
\newcommand{\MDCA}{\texttt{MDCA}\xspace}
\newcommand{\cnns}{\texttt{CNNs}\xspace}
\newcommand{\rnns}{\texttt{RNNs}\xspace}
\newcommand{\CNN}{\texttt{CNN}\xspace}
\newcommand{\ReLU}{\texttt{ReLU}\xspace}
\newcommand{\RNN}{\texttt{RNN}\xspace}
\newcommand{\BiLSTM}{\texttt{BiLSTM}\xspace}
\newcommand{\CTC}{\texttt{CTC}\xspace}
\newcommand{\GRU}{\texttt{GRU}\xspace}
\newcommand{\WRRs}{\texttt{WRRs}\xspace}
\newcommand{\SIFT}{\texttt{SIFT}\xspace}
\newcommand{\SURF}{\texttt{SURF}\xspace}
\newcommand{\BRISK}{\texttt{BRISK}\xspace}
\newcommand{\ORB}{\texttt{ORB}\xspace}
\newcommand{\RANSAC}{\texttt{RANSAC}\xspace}
\newcommand{\MAGSAC}{\texttt{MAGSAC}\xspace}
\newcommand{\ECE}{\texttt{ECE}\xspace}
\newcommand{\SCE}{\texttt{SCE}\xspace}
\newcommand{\AECE}{\texttt{AECE}\xspace}
\newcommand{\etl}{\textit{et al.}\xspace}
\newcommand{\DCA}{\texttt{DCA}\xspace}
\newcommand{\GMM}{\texttt{GMM}\xspace}
\newcommand{\myname}{\texttt{D-MASTER}\xspace}
\newcommand{\inbreast}{\texttt{INBreast}\xspace}
\newcommand{\ddsm}{\texttt{DDSM}\xspace}
\newcommand{\rsna}{\texttt{RSNA-BSD1K}\xspace}
\newcommand{\maes}{\texttt{MAEs}\xspace}
\newcommand{\mrt}{\texttt{MRT}\xspace}
\newcommand{\defdetr}{\texttt{DefDETR}\xspace}
\newcommand{\doad}{\texttt{DOAD}\xspace}
\newcommand{\ssod}{\texttt{SSOD}\xspace}
\newcommand{\ssl}{\texttt{SSL}\xspace}
\newcommand{\mrti}{\texttt{MRTI}\xspace}
\newcommand{\mcm}{\texttt{MCM}\xspace}
\newcommand{\mrm}{\texttt{MRM}\xspace}
\newcommand{\N}{\mathcal{N}}
\newcommand{\G}{\mathcal{G}}
\newcommand{\loss}{\mathcal{L}}
\newcommand{\aA}{\mathcal{a}}
\newcommand{\pP}{\mathcal{p}}
\newcommand{\nN}{\mathcal{n}}
\newcommand{\Dset}{\mathcal{D}}
\newcommand{\X}{\mathcal{X}}
\newcommand{\Y}{\mathcal{Y}}
\newcommand{\yh}{\widehat{y}}
\newcommand{\At}{\widetilde{A}}
\newcommand{\Ct}{\widetilde{C}}
\newcommand{\Lagr}{\mathcal{L}}

%% file: ICLR/sec/0_abstract.tex
\begin{abstract}

Accurate multivariate time-series forecasting is crucial for understanding and mitigating the effects of climate change, as reliable long-horizon predictions support effective monitoring and informed decision-making. Existing neural approaches, ranging from CNNs and RNNs to attention-based Transformers, have achieved notable progress. Yet, they often suffer from two key limitations: difficulty in capturing hierarchical spatiotemporal dependencies and computational inefficiencies when scaling to high-dimensional meteorological data. We propose \fate (Focal-modulated Attention Encoder), a new Transformer architecture tailored for robust multivariate time-series forecasting. \fate introduces a tensorized focal modulation mechanism that enhances spatiotemporal dependency modeling while maintaining scalability. To improve interpretability, we further design dual modulation scores that identify critical environmental features driving the forecasts. Comprehensive experiments on seven diverse real-world datasets, including benchmark energy, traffic, and large-scale climate datasets, demonstrate that \fate consistently surpasses state-of-the-art methods, particularly on long-horizon and high-variability settings. Extensive ablations confirm the generalization ability of \fate across heterogeneous forecasting tasks. To foster reproducibility and future research, we will release the full implementation.

\end{abstract}
 

%% file: ICLR/sec/1_intro.tex
\section{Introduction}
\label{sec:intro}
The Transformer architecture~\citep{Transformer} has become a cornerstone of modern deep learning, driving breakthroughs in natural language processing~\citep{brown2020language, radford2019language, devlin2018bert, CLIP_ICML21}, computer vision~\citep{dosovitskiy2020image, defdetr, yang2022focal}, and large-scale foundation models~\citep{kaplan2020scaling}. Motivated by this success, recent works have applied Transformers to multivariate time-series forecasting, leveraging their ability to model pairwise dependencies and extract multi-level sequence representations~\citep{Autoformer, PatchTST}. However, their effectiveness in this domain remains contested. Notably, simple linear models rooted in classical statistics~\citep{box1968some} have been shown to outperform Transformers in both accuracy and efficiency~\citep{DLinear, das2023long}. At the same time, emerging architectures that explicitly model multivariate correlations~\citep{Crossformer, TSMixer} underscore the limitations of vanilla self-attention for complex time-series dynamics.

We identify three fundamental shortcomings of existing Transformer-based approaches for multivariate forecasting:  
(1) \textit{Permutation-invariant self-attention} fails to capture \textit{temporal order}, leading to weak modeling of \textit{sequential dynamics}.  
(2) \textit{Uniform attention across tokens} not only overlooks the \textit{varying significance of climate variables across spatiotemporal scales}, but also leads to \textit{computational inefficiencies} when scaling to \textit{high-dimensional meteorological data}.  
(3) The architecture lacks an explicit mechanism to model \textit{hierarchical spatiotemporal correlations}, which are crucial for \textit{long-horizon forecasting}.

Unlike FocalNet~\citep{yang2022focal}, which was designed for spatial representation learning in vision tasks, \fate introduces key innovations tailored for multivariate time-series forecasting:  
\begin{itemize}
    \item \textbf{Tensorized Attention Design:} \fate preserves the full 3D tensor structure ($X \in \mathbb{R}^{T \times S \times P}$), maintaining temporal and variable axes explicitly. This enables more effective modeling of long-range dependencies through grouped attention across both time and features.  
    \item \textbf{Focal Grouping for Temporal Blocks:} Instead of spatial grids, \fate dynamically defines \textit{temporal focal groups} that adapt to prediction horizons, allowing the model to capture hierarchical temporal dependencies unique to time-series data.  
    \item \textbf{Cross-axis Modulation:} Focal modulation is extended beyond temporal steps to the variable dimension, enabling rich cross-feature interactions that are absent in FocalNet.  
\end{itemize}

In this way, \fate is not a simple adaptation of FocalNet, but a principled redesign that leverages the structural properties and forecasting demands of multivariate time-series data.

Long-term variations in temperature, precipitation, wind, and other environmental factors define climate change~\citep{barrett2015climate}. These shifts have profound global impacts, threatening sustainability in domains such as food security, public health, and energy systems. For instance, a projected increase of up to $2^\circ$C in global mean temperature this century could severely reduce crop yields. Unlike short-term fluctuations, climate change evolves over decades, driven primarily by greenhouse gas emissions, deforestation, and limited adoption of renewable energy~\citep{latake2015greenhouse}. 
Accurate long-horizon forecasting of such multivariate processes is therefore critical. It enables policymakers and practitioners to assess risks, monitor climate drivers, and design mitigation strategies~\citep{huntingford2019machine}. However, the multidimensional and highly correlated nature of climate data poses significant challenges for existing forecasting models.

\begin{figure}[t]
    \centering
    \begin{subfigure}[b]{0.45\textwidth}
        \centering
        \includegraphics[width=\textwidth]{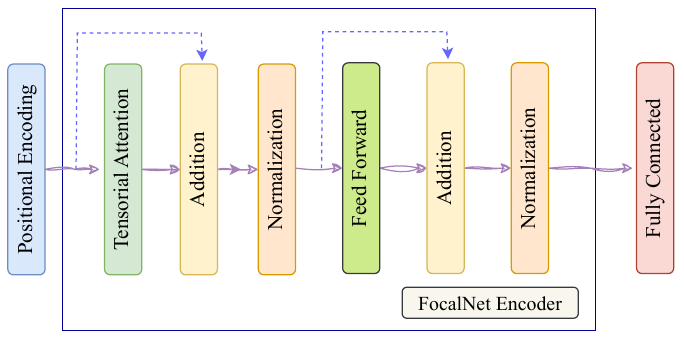}
        \caption{Encoder Architecture of \fate.}
        \label{fig:fate_a}
    \end{subfigure}
    \hfill
    \begin{subfigure}[b]{0.45\textwidth}
        \centering
        \includegraphics[width=\textwidth]{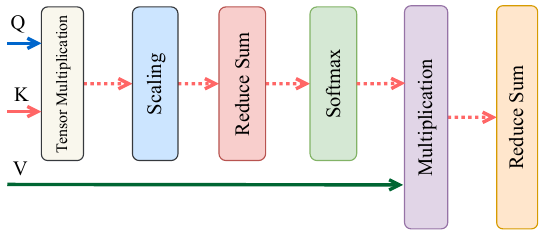}
        \caption{Tensorial Focal Modulation}
        \label{fig:fate_b}
    \end{subfigure}
    \caption{Our proposed architecture consists of two main components: Figure~\ref{fig:fate_a} shows the overall architecture of \fate encoder. The input time series data passes first through positional encoding, and then Tensorial Attention, which incorporates spatial as well as temporal information.  Figure~\ref{fig:fate_b} explains the internal working of the tensorial focal-modulation block. The Query (Q), Key (K), and Value (V) tensors undergo a series of tensor multiplication, scaling, reduction, and softmax operations to create attention maps. These maps are then used by the model to determine which regions of the inputs are more significant.}  
    \label{fig1}
\end{figure}

To address these challenges, we propose \fate, a novel Transformer that 
(1) introduces tensorized focal modulation for explicit spatiotemporal correlation modeling, 
(2) employs dual modulation scores to enhance interpretability, and 
(3) adaptively emphasizes relevant tokens via selective attention.
We evaluate \fate across seven diverse real-world datasets and demonstrate that it consistently outperforms state-of-the-art methods, particularly on long-horizon and high-dimensional climate datasets. Extensive ablation studies further confirm that \fate generalizes effectively across broader multivariate forecasting tasks.

\noindent\textbf{Contributions.} The main contributions of this work are threefold:  
\begin{itemize}
    \item We introduce \fate, a Transformer architecture with a novel focal-modulation mechanism that preserves 3D tensor structure ($T \times S \times P$) for multivariate time-series forecasting.  
    \item We design \textit{dual modulation scores} that improve both predictive performance and interpretability by identifying critical temporal and variable dependencies.  
    \item We achieve new state-of-the-art results on seven benchmark datasets, including accuracy gains of $13.3\%$, $9.1\%$, and $10.1\%$ on \texttt{ETTm2}~\citep{haoyietal-informer-2021}, \texttt{Weather5k}~\citep{han2024weather5k}, and \texttt{LargeST}~\citep{liu2023largest}, respectively, with strong improvements across all other datasets.
\end{itemize}

%% file: ICLR/sec/2_background_and_related_work.tex
\section{Related Work}
\label{sec:related_work}

\textbf{Transformers for Time Series Forecasting.}  
Transformer architectures~\citep{Transformer} have achieved remarkable success across NLP~\citep{bert, brown2020language, radford2019language}, computer vision~\citep{vit, beit, mae}, and speech~\citep{wav2vec2, hubert} due to their scalability and effective sequence modeling. Vision Transformers (ViTs) divide images into patches to preserve local semantic information~\citep{vit, geiger2013vision, li2020spatial}, while NLP models like BERT~\citep{devlin2018bert} leverage subword tokenization for contextual dependencies.  
Inspired by these successes, Transformer variants have been widely adapted for time-series forecasting~\citep{spacetimeformer, PatchTST}. Early models, such as LogTrans~\citep{logtrans} and Informer~\citep{informer}, addressed computational inefficiencies via sparse attention. Autoformer~\citep{Autoformer} introduced decomposition-based inductive biases, FEDformer~\citep{fedformer} employed Fourier-enhanced blocks for seasonal modeling, Pyraformer~\citep{pyraformer} added pyramidal attention for multi-scale dependencies, and Triformer~\citep{triformer} proposed pseudo-timestamp-based patch attention.  
Despite these advances, many Transformer forecasters still rely on point-wise or handcrafted attention, limiting their ability to capture semantic relationships across patches or dimensions~\citep{sakaridis2018semantic, ashish2017attention, zhu2023patch}. For example, Autoformer’s fixed auto-correlation modules may fail to generalize, and Triformer does not treat patches as first-class units nor model internal semantics. TimeMixer++~\citep{wang2024timemixer++} advances multi-scale, multi-resolution forecasting by converting time series into 2D time images (via Multi-Resolution Time Imaging, \mrti) and separating seasonal/trend components in latent space using dual-axis attention, followed by hierarchical Multi-Scale Mixing (\mcm) and Multi-Resolution Mixing (\mrm). This allows parallel modeling of concurrent temporal contexts (daily, weekly, seasonal), improving forecasting, classification, and anomaly detection.  
TimeTensor~\citep{liang2024timetensor} generalizes linear attention to 3D tensor inputs via Kronecker decomposition, improving efficiency while retaining the standard attention paradigm. In contrast, \fate introduces \textit{tensorized focal modulation}, explicitly preserving 3D spatiotemporal structure, enabling hierarchical and localized context aggregation, and jointly modeling long- and short-range dependencies. This represents a novel architectural strategy distinct from previous tensorized attention mechanisms.

\textbf{Self-supervised and Representation Learning.}  
Transformer adaptations for time series can be categorized into four directions~\citep{nlpsurvey}: (i) attention-level modifications for efficiency, (ii) adaptations for stationarity and signal processing, (iii) architectural changes capturing cross-variate and temporal dependencies, and (iv) novel tensor-based designs. Most methods focus on the first three, while few explore fundamental tensor-based redesigns.  
Self-supervised learning (SSL) has also gained traction for time-series representation learning. Methods such as TNC~\citep{tnc}, TS2Vec~\citep{ts2vec}, and BTSF~\citep{btsf} learn rich representations without supervision, whereas Transformer-based SSL models like TST~\citep{tst} and TS-TCC~\citep{ts-tcc} remain underexplored for capturing complex temporal and cross-variate dependencies. \fate’s tensorized focal modulation inherently supports richer hierarchical representations, bridging this gap by jointly modeling time, feature, and spatial dimensions.  
\textit{Focal Modulated Tensorized Encoder} introduces a novel tensorized focal modulation mechanism tailored for multivariate time-series forecasting. It preserves the input’s 3D tensor structure ($T \times S \times P$), enables hierarchical spatiotemporal correlation modeling, and applies \textit{tensorized attention design}, \textit{temporal focal grouping}, and \textit{cross-axis modulation}. Unlike prior work, \fate balances efficiency with semantic richness and provides a principled framework for long- and short-range dependency modeling in high-dimensional time series.

%% file: ICLR/sec/3_purpose_meth.tex
\section{Proposed Methodology}
In this section, we present \fate, a \textit{Focal Modulated Tensorized Encoder Transformer} designed for multivariate time-series forecasting. The architecture preserves the full 3D structure of the input tensor to jointly model temporal, spatial (station-wise), and feature dimensions. Central to \fate are tensorized focal modulation mechanisms that efficiently capture hierarchical temporal patterns, cross-station interactions, and feature dependencies, while providing interpretable modulation scores that highlight the contribution of each station and attention head. The following subsections detail the encoder design, the tensorial focal modulation computations, and the aggregation strategy for interpretable predictions.

\subsection{Multi-Dimensional Tensored FocalNet Encoder}
We extend the \texttt{FocalNet} Transformer~\citep{yang2022focal} to propose the \textit{Tensorized Focal Encoder Transformer}, specifically designed to capture complex patterns in multi-dimensional time-series data. Our model operates on climate parameters organized as a 3D tensor \(X \in \mathbb{R}^{T \times S \times P}\), where \(T\) denotes the temporal dimension, \(S\) indexes different stations, and \(P\) represents diverse climate parameters (e.g., temperature, humidity, wind speed). The full 3D structure preserves variable–time step relationships and supports parallel yet separate attention across temporal and feature dimensions.  


The architecture is encoder-only, as illustrated in Figure~\ref{fig1}, and comprises: (i) a positional encoding layer, (ii) a tensorial focal modulation encoder layer, and (iii) a linearly activated fully-connected layer. Each encoder layer integrates tensorial modulation (Sections 3.2 and 3.3) followed by a residual connection and normalization. A densely connected \texttt{FFN}, consisting of two linear transformations with \texttt{ReLU} activation, follows the modulation layer, and is again succeeded by residual connection and normalization, consistent with~\citep{yang2022focal}.

\subsection{Tensorial Focal Modulation}

To encode temporal hierarchies, we apply a constant positional encoding~\citep{yang2022focal} along the time axis \(T\) and parameter axis \(P\):

\begin{equation}
    \text{PE}(\text{pos}, 2i) = \sin\left(\frac{\text{pos}}{10000 \cdot 2^i / P}\right),
\label{eq1}
\end{equation}

where \(\text{pos}\) indexes time and \(i\) indexes parameters; the station axis \(S\) transmits the encoded values.

Focal modulation replaces pairwise attention with hierarchical context aggregation~\citep{yang2022focal}, offering three key benefits: (i) improved computational efficiency, (ii) preservation of locality biases, and (iii) non-quadratic long-range dependency modeling. For multivariate time series, \fate leverages this through: (1) nested focal windows that hierarchically aggregate temporal information, and (2) dynamic contextual gating that adapts to input distributions, outperforming fixed receptive fields or conventional attention kernels.

We formalize tensor slices as follows: for a tensor \(N \in \mathbb{R}^{X \times Y \times Z}\), \((N_{y,z})_x \in \mathbb{R}^{Y \times Z}\) denotes the \(x\)-slice, and \((N_z)_{x,y} \in \mathbb{R}^{Z}\) denotes the \(x, y\)-slice. Lowercase letters indicate slice sizes.

Tensorial focal modulation operates on \(X \in \mathbb{R}^{T \times S \times P}\). We first compute 3D Query (\(Q\)), Key (\(K\)), and Value (\(V\)) tensors, \(Q,K,V \in \mathbb{R}^{T \times S \times H}\), via element-wise multiplication with learnable weight tensors \(W^Q,W^K,W^V \in \mathbb{R}^{S \times F \times H}\):

\begin{equation}
\begin{aligned}
(Q_h)_{t,s} &= (X_p)_{t,s} \times (W^Q)_{p,h,s},\\
(K_h)_{t,s} &= (X_p)_{t,s} \times (W^K)_{p,h,s},\\
(V_h)_{t,s} &= (X_p)_{t,s} \times (W^V)_{p,h,s}, \quad \forall t = 1..T, s = 1..S.
\end{aligned}
\label{eq2}
\end{equation}


Next, we compute the multiplicative interaction across time steps:

\begin{equation}
(\widetilde{R}_{s,s^l})_{t,t^l} = (Q_{s,h})_t \times ((K_{s^l,h})_{t^l})^T, \quad R = \frac{1}{\sqrt{H}} \sum_{s^l=1}^{S} (\widetilde{R}_{t,t^l,s})_{s^l},
\label{eq3}
\end{equation}

followed by a softmax across the station dimension to obtain attention weights \(\widetilde{A} \in \mathbb{R}^{T \times T^l \times S}\):

\begin{equation}
(\widetilde{A}_s)_{t,t^l} = \text{Softmax}\left((R_{t,t^l,s})_s\right), \quad \forall t, t^l = 1..T.
\label{eq4}
\end{equation}

Finally, the output \(Z \in \mathbb{R}^{T \times C \times D}\) is computed by broadcasting \((\widetilde{A}_s)_{t,t'}\) to match the shape of \((V_{s,d})_{t'}\) and summing over the temporal dimension:

\begin{equation}
(Z_{s,d})_{t} = \sum_{t'=1}^{T} \text{broadcast}((\widetilde{A}_s)_{t,t'}) \circ (V_{s,d})_{t'}, \quad \forall t = 1..T.
\label{eq5}
\end{equation}

\subsection{Focal Modulation Aggregation} 

Modulation weights have been widely used for feature selection and interpretability~\citep{wiegreffe2019attention}. In \fate, the focal modulation tensors \(\widetilde{A}\) (Eq.~\ref{eq4}) serve to provide interpretable insights into model predictions.

To quantify the relationship between attention heads and cities (stations), we compute \textit{head-wise focal modulation scores}:

\begin{equation} \label{eq:s_hc} 
N\widetilde{A}_{s}^{h} = \sum_{t=1}^{T} \sum_{t'=1}^{T'} A_{t,t',c}^h, \quad \forall h = 1..H, \; c = 1..C.
\end{equation}

We then aggregate across all heads to obtain \textit{city-wise modulation scores}, reflecting the overall contribution of each city to the prediction:

\begin{equation} \label{eq:s_c} 
N\widetilde{A}_{s} = \sum_{h=1}^{H} N\widetilde{A}_{s}^{h}, \quad \forall c = 1..C.
\end{equation}    

This aggregation completes the tensorial focal modulation process, explicitly linking attention heads to cities and highlighting the importance of each city in driving the model’s forecasts.

%% file: ICLR/sec/4_experiments.tex
\section{Experiments}
To rigorously evaluate \fate, we conduct extensive experiments on seven diverse real-world datasets spanning environmental and infrastructural domains, comparing against 17 state-of-the-art baselines including Transformer-, RNN/CNN-, Linear-, and spatial-temporal models. We analyze predictive performance across short- and long-horizon forecasts using standard metrics (\mae, \mse), benchmark computational and memory efficiency, and provide interpretability through focal modulation visualization. These studies demonstrate \fate's superior accuracy, robustness, and capacity to model multi-scale temporal and spatiotemporal dependencies.

\subsection{Datasets}
We evaluate \fate on seven diverse real-world multivariate time-series datasets, encompassing both environmental (Weather5k, USA-Canada, Europe) and infrastructural (ETTh1, ETTm2, Traffic, LargeST) domains.

\textbf{ETTh1}~\cite{haoyietal-informer-2021} and \textbf{ETTm2}~\cite{haoyietal-informer-2021} are electricity transformer datasets at hourly and minute resolutions, respectively, capturing seasonal and trend-driven consumption patterns. \textbf{Traffic}~\cite{traffic_flow_forecasting_608} consists of road occupancy rates from multiple sensors, serving as a standard benchmark for traffic flow prediction. \textbf{Weather5k}~\cite{han2024weather5k} is a large-scale dataset with 10 years of hourly measurements from 5,672 weather stations worldwide, including temperature, humidity, wind speed, and other climate parameters. \textbf{USA-Canada}~\cite{cisl_rda_dsd472000} contains hourly meteorological data from 30 cities (Oct 2012–Nov 2017), enriched with spatial coordinates and temporal features such as hour and day-of-year. The \textbf{Europe} dataset~\cite{huber2022_weather} spans 18 European cities (May 2005–Apr 2020), with normalized temporal and meteorological features; the test split covers 2017–2020, and the training/validation span 2005–2017. Finally, \textbf{LargeST}~\cite{liu2023largest} provides traffic data from 8,600 sensors in California over 5 years, including rich sensor metadata for enhanced interpretability.  

Across all datasets, \fate consistently outperforms baselines—including Transformer~\cite{vaswani2017attention, yang2022focal}, 3D-\cnn~\cite{mehrkanoon2019deep}, \lstm~\cite{hochreiter1997lstm}, and ConvLSTM~\cite{shi2015convolutional}—achieving the lowest Mean Absolute Error (\mae) and Mean Squared Error (\mse), particularly on long-horizon and high-dimensional climate datasets.

\subsection{Additional Implementation Details}

\textbf{Computational and Memory Requirements.}  
We use a fixed 30-day input window; for climate datasets, we consider 7 meteorological features, while feature selection for other datasets follows the original data schema. Experiments were conducted on an NVIDIA A100 GPU with 40GB VRAM. Optimizers were selected per architecture following prior best practices.  

We analyze \fate’s computational complexity and provide empirical runtime benchmarks against Transformer and CNN-based baselines. While tensorized focal modulation introduces moderate overhead compared to standard Transformers, the performance gains in long-horizon forecasting justify this cost. Preserving the 3D tensor increases memory complexity due to grouped modulation, but efficient projections keep runtime and GPU usage comparable to baseline Transformers.
  
\input{Tables/hyperparameters}
\textbf{Hyperparameters.}  
Table~\ref{tab:hyperparams} details all training hyperparameters. Multi-head attention is used in both \fate and Transformer models, with \fate employing four focal levels and eight attention heads to capture hierarchical temporal dependencies. 3D-\cnn~\cite{mehrkanoon2019deep} and Conv\lstm~\cite{shi2015convolutional} models use convolutional layers with kernel sizes tuned for spatiotemporal patterns. \lstm~\cite{hochreiter1997lstm} and Conv\lstm models employ recurrent units with hidden dimensions optimized for sequential modeling. Scheduled learning rate decay is applied in \fate and Transformer models, while 3D-\cnn, \lstm, and Conv\lstm use fixed rates. Batch sizes are scaled for memory efficiency and stable training.

\subsection{Forecasting Results}
\input{Tables/results_ETTM}
We evaluate \fate across diverse real-world datasets and benchmark it against 17 state-of-the-art models spanning four categories: 
(1) \textit{Transformer-based}: iTransformer~\cite{nie2024itransformer}, Autoformer~\cite{wu2021autoformer}, etc.; 
(2) \textit{RNN/CNN-based}: LSTM~\cite{hochreiter1997lstm}, ConvLSTM~\cite{shi2015convolutional}, 3D-\cnn~\cite{mehrkanoon2019deep}; 
(3) \textit{Linear-based}: DLinear~\cite{zeng2023dlinear}, TiDE~\cite{das2023tide}; 
(4) \textit{Spatial-temporal} (LargeST dataset): DGCRN~\cite{li2023dynamic}, D2STGNN~\cite{shao2022decoupled}.

\noindent \textbf{Continental-scale Forecasting.}  
On USA-Canada and Europe datasets, we evaluate 4–16 hour forecasts using \mae and \mse (Table~\ref{tab4}). \fate consistently outperforms all baselines, including robust Transformers and linear models. For example, in Vancouver, \fate reduces \mae and \mse by up to 15.9\% and 24.9\%, respectively, over the best baseline. The Europe dataset exhibits similar trends, highlighting \fate's robustness and ability to model long-horizon temporal dynamics effectively.

\noindent \textbf{Large-Scale Spatiotemporal Forecasting.}  
On the LargeST dataset (Table~\ref{tab:lagest}), \fate achieves the lowest \mae and \mse (0.160 and 0.255), surpassing D2STGNN by 10.1\% and 13.6\%, respectively. These results demonstrate \fate's capacity to capture intricate spatiotemporal dependencies in large-scale traffic data, making it highly suitable for real-world forecasting applications.

\input{Tables/results_combined}

\noindent \textbf{Benchmark Dataset Evaluation.}  
Across four standard multivariate time series datasets (ETTh1, ETTm2, Traffic, Weather5k; Table~\ref{tab:combined_4}), \fate consistently achieves state-of-the-art or competitive performance. Notably:  
- ETTh1: 4.3\% lower \mae, capturing fine-grained temporal patterns.  
- Traffic: compared to PatchTST, \mae is slightly higher, but \mse decreases by 3\%.  
- Weather5k: 9.1\% \mae and 12.3\% \mse improvements over CI-TSMixer, demonstrating robustness to high-dimensional noise.  
- ETTm2: 13.3\% \mae and 7.9\% \mse improvements, confirming generalizability across diverse datasets.
These results collectively validate \fate’s ability to model multi-scale temporal and spatial dependencies, yielding accurate and stable forecasts across both regional and large-scale datasets.
\input{Tables/combined_table}

%% file: Tables/hyperparameters.tex

\begin{wraptable}{r}{0.5\textwidth}  
\centering
\small
\vspace{-1em}
\caption{Hyperparameters used for all the models. All hyperparameters were selected using 5-fold cross-validation. Tuning was done independently on each dataset to avoid overfitting or unfair transfer of settings.}
\vspace{1em}
\setlength{\tabcolsep}{6pt}
\begin{adjustbox}{max width=\linewidth}
\begin{tabular}{l|c|c|c|c|c}
\toprule
\textbf{Hyper-parameter} & \textbf{FATE} & \textbf{Transformer} & \textbf{3D CNN} & \textbf{LSTM} & \textbf{ConvLSTM} \\
\midrule
Focal Levels   & 4  & 3  & -   & -   & -   \\
Layer Number   & 1  & 1  & -   & 1   & 3   \\
Head           & 8  & 1  & -   & -   & -   \\
Key Dim        & 32 & 32 & -   & -   & -   \\
Dense Units    & 64 & 64 & 128 & -   & -   \\
Filters        & -  & -  & 10  & -   & 16  \\
Kernel Size    & -  & -  & 4   & -   & 13  \\
Hidden Units   & -  & -  & -   & 128 & -   \\
Learning Rate  & Schedule & Schedule & $10^{-4}$ & $10^{-4}$ & $10^{-4}$ \\
Batch Size     & 64 & 32 & 128 & 256 & 128 \\
\bottomrule
\end{tabular}
\end{adjustbox}
\vspace{1em}
\label{tab:hyperparams}
\end{wraptable}

%% file: Tables/results_ETTM.tex
\begin{wraptable}{r}{0.5\textwidth}
    \centering
    \small
    \vspace{-0.8em}
    \caption{Comparison of model performance on \textit{LargeST} dataset. The best performing model is shown in \textbf{bold} and the second best in \textcolor{red}{red} for clarity.}
    \vspace{1em}
    \begin{adjustbox}{max width=\linewidth}
    \begin{tabular}{lcc}
        \toprule
        \textbf{Model} & \textbf{MAE} & \textbf{MSE} \\
        \midrule
        LSTM~\cite{hochreiter1997lstm} & 0.266 & 0.417 \\
        DRCNN~\cite{9271925} & 0.213 & 0.333 \\
        STNN~\cite{9763814} & 0.186 & 0.311 \\
        STGODE~\cite{fang2021spatial} & 0.195 & 0.335 \\
        DGCRN~\cite{li2023dynamic} & 0.180 & 0.300 \\
        D2STGNN~\cite{shao2022decoupled} & \textcolor{red}{0.178} & \textcolor{red}{0.295} \\
        \midrule
        FATE (Ours) & \textbf{0.160} & \textbf{0.255} \\
        \bottomrule
    \end{tabular} 
    \end{adjustbox}
    \vspace{1em}
    \label{tab:lagest}
\end{wraptable}

%% file: Tables/results_combined.tex
\begin{table*}[!htb]
    \centering
    \caption{\small{The test results for temperature prediction, evaluated using the Mean Absolute Error (\mae) and Mean Squared Error (\mse), were obtained for the \textbf{USA-Canada} and \textbf{Europe} datasets. The best-performing results are highlighted in \textbf{bold}, while the second-best are marked in \textcolor{red}{red} for clarity.}}
    \vspace{-0.5em}
    \begin{adjustbox}{max width=\textwidth}
        \begin{tabular}{l l | cccc | cccc || l l | ccc | ccc}
        
            \toprule
            
            \multirow{2}{*}{Station} & \multirow{2}{*}{Model} & \multicolumn{4}{c|}{\mae} & \multicolumn{4}{c||}{\mse} & 
            
            \multirow{2}{*}{Station} & \multirow{2}{*}{Model} & \multicolumn{3}{c|}{\mae} & \multicolumn{3}{c}{\mse} \\
            
            & & {4 hrs} & {8 hrs} & {12 hrs} & {16 hrs} & {4 hrs} & {8 hrs} & {12 hrs} & {16 hrs} &
            & & 3 days & 5 days & 7 days & 3 days & 5 days & 7 days \\
            
            \midrule
            
            \multirow{15}{*}{\textbf{\textbf{Vancouver}}}

            & Transformer~\cite{vaswani2017attention}   & 1.238 & 1.858 & 1.987 & 2.146 & 2.566 & 5.787 & 6.617 & 7.748 &
            \multirow{15}{*}{\textbf{Barcelona}} 
            & Transformer~\cite{vaswani2017attention}   & 2.608 & 2.901 & 3.347 & 11.702 & 14.660 & 15.926 \\
            & 3D \cnn~\cite{MEHRKANOON_RepresLearning}                           & 1.499 & 1.896 & 2.131 & 2.329 & 3.704 & 5.950 & 7.455 & 8.879 &
            & 3D \cnn~\cite{MEHRKANOON_RepresLearning}                           & 2.502 & 3.015 & 3.059 & 10.73 & 13.654 & 15.740 \\
            & \lstm~\cite{hochreiter1997lstm}           & 1.311 & 1.834 & 2.039 & 2.210 & 2.917 & 5.712 & 6.970 & 8.237 &
            & \lstm~\cite{hochreiter1997lstm}           & \textcolor{red}{2.303} & 2.801 & \textcolor{red}{2.931} & \textcolor{red}{9.354} & 11.328 & 14.931 \\
            & Conv\lstm~\cite{shi2015convolutional}     & 1.338 & 1.829 & 1.992 & 2.194 & 2.967 & 5.553 & 6.571 & 7.990 &
            & Conv\lstm~\cite{shi2015convolutional}     & 2.759 & {2.787}  & 2.948 & 12.882 & {12.272} & {14.920} \\ 

            & Autoformer~\cite{wu2021autoformer}        &  1.258   &1.982 & 2.682&2.695 & 2.578&4.598 & 6.395&6.087 &
            & Autoformer~\cite{wu2021autoformer}        &2.798     &2.878 &3.212 &12.489 &12.976 &15.345 \\
            & SCINet~\cite{liu2022scinet}               &1.458    &2.905 &1.890 &1.870 &2.880 &5.456 &6.873 &8.293 &
            & SCINet~\cite{liu2022scinet}               &2.902     &2.789 &3.404 &12.213 &13.643 &15.895 \\ 
            & FEDformer~\cite{zhou2022fedformer}        &1.590        &1.563 &1.992 &2.809 &3.556 &5.679 &6.163 &6.946 &
            & FEDformer~\cite{zhou2022fedformer}        &2.709        &2.778 &3.112 &11.678 & 13.234&15.543 \\ 
            & Stationary~\cite{liu2022stationary}       &1.354     &\textcolor{red}{1.430} &2.058 &2.890 &3.050 &4.987 &6.201 &7.845 &
            & Stationary~\cite{liu2022stationary}       &2.765     &2.987 &3.641 &12.975 &12.075 &14.887\\
            & RLinear~\cite{li2023rlinear}              &1.673       &1.256 &1.890 &2.450 &2.990 &4.678 &5.987 &6.289 &
            & RLinear~\cite{li2023rlinear}              &2.834       &3.543 &3.342 &11.897 &12.675 &15.967 \\ 
            & PatchTST~\cite{li2023patchtst}            &1.568        &1.789 &1.640 &2.180 &1.764 &4.234 &5.239 &3.923&
            & PatchTST~\cite{li2023patchtst}            &2.623        &3.234 &3.375 &12.456 &11.907 &15.325\\ 
            & Crossformer~\cite{zhang2023crossformer}   &1.456        &1.590 &1.678 &1.990 &1.678 &3.989 &4.786 &5.257 &
            & Crossformer~\cite{zhang2023crossformer}   &2.854        &2.878 &3.123 &12.654 &12.985 &15.564 \\ 
            & TiDE~\cite{das2023tide}                   &1.555       &1.728 &1.430 &1.789 &2.278 &3.278 &3.987 &4.890 &
            & TiDE~\cite{das2023tide}                   &2.542     &2.690 &3.078 &12.267 &12.754 &14.243 \\ 
            & TimesNet~\cite{wu2023timesnet}            &1.145        &1.567 &1.678 &1.890 &2.789 &2.908 &3.678 &3.980 &
            & TimesNet~\cite{wu2023timesnet}            &2.876        &2.879 &3.321 &11.654 &11.754 &14.675 \\ 
            & DLinear~\cite{zeng2023dlinear}            &1.134        &1.556 &1.567 &1.567 &1.890 &2.465 &2.967 &3.653 &
            & DLinear~\cite{zeng2023dlinear}            &2.567      &3.165 &3.145 &11.687 &11.946 &13.990 \\ 
            & iTransformer~\cite{nie2024itransformer}   &\textcolor{red}{1.123}       &1.487 &\textcolor{red}{1.435} &\textcolor{red}{1.345} &\textcolor{red}{1.670} &\textcolor{red}{1.910} &\textcolor{red}{2.456} &\textcolor{red}{2.847} &
            & iTransformer~\cite{nie2024itransformer}   &2.680        &2.989 &3.076 &10.456 &11.896 &13.696 \\
            \cmidrule(lr){2-18}
            & \fate \textbf{ (Ours) }                             & \textbf{1.021} & \textbf{1.217} & \textbf{1.346} & \textbf{1.131} & \textbf{1.464} & \textbf{1.660} & \textbf{1.844} & \textbf{2.238} &
            & \fate \textbf{ (Ours) }                             & \textbf{2.174} & \textbf{2.665} & \textbf{2.695} & \textbf{8.515} & \textbf{10.914} & \textbf{13.523} \\
           
            \midrule
            
            \multirow{15}{*}{\textbf{New York}}

            & Transformer~\cite{vaswani2017attention}   & 1.426 & 2.043 & 2.271 & 2.489 & 3.836 & 7.533 & 9.268 & 10.978 &
            \multirow{15}{*}{\textbf{Maastricht}}
            & Transformer~\cite{vaswani2017attention}   & 4.770 & 5.293 & 5.649 & 30.891 & 43.283 & 50.678 \\
            & 3D \cnn~\cite{MEHRKANOON_RepresLearning}                           & 1.835 & 2.316 & 2.833 & 2.673 & 5.587 & 9.159 & 13.468 & 11.964 &
            & 3D \cnn~\cite{MEHRKANOON_RepresLearning}                           & 4.276 & 5.078& 5.609 & 28.823& 40.531& 49.410\\
            & \lstm~\cite{hochreiter1997lstm}           & 1.596 & 2.126 & 2.325 & 2.507 & 4.724 & 8.103 & 9.749 & 10.985 &
            & \lstm~\cite{hochreiter1997lstm}           & 3.982 & 5.036& 5.373 & \textcolor{red}{24.860} & 39.484 & 46.590\\
            & Conv\lstm~\cite{shi2015convolutional}     & 1.394 & 2.134 & 2.419 &\textcolor{red}{ 2.104} & 4.949 & 7.790 & 9.257 & 10.341 &
            & Conv\lstm~\cite{shi2015convolutional}     & 4.578& \textcolor{red}{4.863}  & \textcolor{red}{5.322}& 32.699& 39.819 & 43.288  \\
            
             & Autoformer~\cite{wu2021autoformer}        &1.756    &1.981 &2.587 &2.446 &4.436 &7.234 &10.457 &9.357 &
            & Autoformer~\cite{wu2021autoformer}        &4.896     &5.987 &5.670 &32.969 &39.563 &48.939 \\
            & SCINet~\cite{liu2022scinet}               &1.940     &1.879 &2.859 &2.976 &4.876 &6.905 &12.755 &10.345 &
            & SCINet~\cite{liu2022scinet}               &4.886     &5.109 &5.348 &33.123 &40.909 &47.834 \\ 
            & FEDformer~\cite{zhou2022fedformer}        &1.650        &\textcolor{red}{1.809} &2.865 &2.768 &4.345 &6.469 &12.657 &9.235 &
            & FEDformer~\cite{zhou2022fedformer}        &4.609        &5.689 &5.456 &32.689 &41.549 &45.834 \\ 
            & Stationary~\cite{liu2022stationary}       &1.903     &1.980 &2.786 &2.567 &3.957 &7.458 &11.466 &9.587 &
            & Stationary~\cite{liu2022stationary}       &4.679     &5.786 &5.940 &32.569 &40.457 &47.394 \\
            & RLinear~\cite{li2023rlinear}              &1.455       &1.912 &2.532 &2.545 &4.768 &7.548 &10.567 &10.344 &
            & RLinear~\cite{li2023rlinear}              &4.798       &5.079 &5.749 &32.564 &41.348 &50.576 \\ 
            & PatchTST~\cite{li2023patchtst}            &1.465        &1.893 &2.230 &2.443 &4.534 &5.990 &13.565 &10.497 &
            & PatchTST~\cite{li2023patchtst}            &4.765        &5.768 &5.088 &31.455 &39.457 &49.785\\ 
            & Crossformer~\cite{zhang2023crossformer}   &2.124        &2.498 &2.432 &2.234 &4.786 &6.935 &11.356 &9.346 &
            & Crossformer~\cite{zhang2023crossformer}   &4.56        &5.698 &5.678 &31.455 &41.694 &46.876 \\ 
            & TiDE~\cite{das2023tide}                   &1.967       &2.231 &2.241 &2.948 &3.654 &7.345 &10.549 &9.438 &
            & TiDE~\cite{das2023tide}                   &4.969       &5.345 &5.543 &31.289 &40.694 &48.567 \\ 
            & TimesNet~\cite{wu2023timesnet}            &1.567        &1.890 &2.532 &2.468 &3.234 &7.095 &9.657 &10.348 &
            & TimesNet~\cite{wu2023timesnet}            &4.579        &5.234 &5.432 &30.234 &41.457 &50.345 \\ 
            & DLinear~\cite{zeng2023dlinear}            &1.563        &2.086 &\textcolor{red}{2.124} &2.983 &3.767 &6.455 &9.378 &9.347 &
            & DLinear~\cite{zeng2023dlinear}            &4.998       &5.234 &5.876 &30.457 &40.345 &49.566 \\ 
            & iTransformer~\cite{nie2024itransformer}   &\textcolor{red}{1.274}        &1.908 &2.343 &2.435 &\textcolor{red}{3.555} &\textcolor{red}{5.839} &\textcolor{red}{7.994} & \textcolor{red}{8.904}&
            & iTransformer~\cite{nie2024itransformer}   &4.458       &5.343 &5.765 &30.578 &\textcolor{red}{39.457} &\textcolor{red}{43.456} \\
            \cmidrule(lr){2-18}
            & \fate \textbf{ (Ours) }                             & \textbf{0.982} & \textbf{1.689} & \textbf{1.974} & \textbf{1.995} & \textbf{3.180} & \textbf{5.296} & \textbf{6.677} & \textbf{8.193} &
            & \fate \textbf{ (Ours) }                             & \textbf{4.164} & \textbf{4.410} & \textbf{4.940} & \textbf{21.458} & \textbf{35.501} & \textbf{39.707} \\
           
            \midrule
            
            \multirow{15}{*}{\textbf{Los Angeles}}

            & Transformer~\cite{vaswani2017attention}   & 1.426 & 2.043 & 2.271 & 2.489 & 3.836 & 7.533 & 9.268 & 10.978 &
            \multirow{15}{*}{\textbf{Munich}}
            & Transformer~\cite{vaswani2017attention}   & 4.136 & 5.286& 5.275& 23.954& 39.057& 43.526\\
            & 3D \cnn~\cite{MEHRKANOON_RepresLearning}                           & 1.835 & 2.316 & 2.833 & 2.673 & 5.587 & 9.159 & 13.467 & 11.968 &
            & 3D \cnn~\cite{MEHRKANOON_RepresLearning}                           & 3.931 & 5.049 & 5.262& 24.870& 39.578& 43.507\\
            & \lstm~\cite{hochreiter1997lstm}           & 1.296 & 2.026 & 2.325 & 2.207 & 4.724 & 8.403 & 9.749 & 10.983 &
            & \lstm~\cite{hochreiter1997lstm}           & {3.551} & {4.730} & 5.189 & \textcolor{red}{20.235} & \textcolor{red}{34.021} & 42.733\\
            & Conv\lstm~\cite{shi2015convolutional}     & 1.594 & 2.134 & 2.419 & 2.704 & 4.949 & 7.790 & 8.457 & 12.342 &
            & Conv\lstm~\cite{shi2015convolutional}     & 3.974& 4.830  & \textcolor{red}{5.023}& 22.484& 35.401& \textcolor{red}{37.767} \\ 
            
             & Autoformer~\cite{wu2021autoformer}        &1.645     &2.457 &2.856 &2.980 &3.886 &9.203 &13.124 &12.588 &
            & Autoformer~\cite{wu2021autoformer}        &3.958     &5.890 &5.456 &22.467 &37.347 &40.458 \\
            & SCINet~\cite{liu2022scinet}               &1.458     &2.346  &2.456 &2.608 &3.508 &9.134 &12.244 &12.458 &
            & SCINet~\cite{liu2022scinet}               &4.545     &5.461 &5.546 &20.567 &40.890 &44.102 \\ 
            & FEDformer~\cite{zhou2022fedformer}        &1.748        &2.479 &2.567 &2.647 &3.680 &7.904 &13.598 &12.453 &
            & FEDformer~\cite{zhou2022fedformer}        &3.957        &5.563 &5.986 &23.467 &39.834 &42.549 \\ 
            & Stationary~\cite{liu2022stationary}       &1.983     &2.986 &2.096 &2.678 &3.976 &8.348 &12.548 &11.579 &
            & Stationary~\cite{liu2022stationary}       &3.589     &5.970 &5.446 &22.366 &38.787 &43.124 \\
            & RLinear~\cite{li2023rlinear}              &1.849       &2.228 &2.345 &2.956 &3.578 &8.438 &11.959 &11.345 &
            & RLinear~\cite{li2023rlinear}              &3.335       &5.348 &5.785 &20.456 &39.456 &42.957 \\ 
            & PatchTST~\cite{li2023patchtst}            &1.648        &2.562 &2.956 &2.907 &3.877 &7.348 &11.345 &10.397 &
            & PatchTST~\cite{li2023patchtst}            &3.595        &5.795 &5.679 &21.458 &40.683 &42.458\\ 
            & Crossformer~\cite{zhang2023crossformer}   &1.843        &2.875 &2.645 &2.845 &3.689 &6.937 &10.543 &10.458 &
            & Crossformer~\cite{zhang2023crossformer}   &3.579        &5.675 &5.685 &23.546 &40.348 &44.939 \\ 
            & TiDE~\cite{das2023tide}                   &1.937       &2.780 &2.454 &2.689 &\textcolor{red}{3.273} &6.348 &9.434 &10.439 &
            & TiDE~\cite{das2023tide}                   &3.584       &5.235 &5.436 &23.754 &39.457 &44.345 \\ 
            & TimesNet~\cite{wu2023timesnet}            &1.893        &2.549 &2.644 &2.997 &3.679 &6.438 &8.934 &9.948 &
            & TimesNet~\cite{wu2023timesnet}            &3.545        &5.344 &5.543 &23.456 &38.458 &42.589 \\ 
            & DLinear~\cite{zeng2023dlinear}            &1.457        &2.456 &2.344 &2.578 &3.879 &\textcolor{red}{6.349} &\textcolor{red}{8.282} &9.348 &
            & DLinear~\cite{zeng2023dlinear}            &3.565        &5.234 &5.567 &22.546 &37.459 &43.548 \\ 
            & iTransformer~\cite{nie2024itransformer}   &\textcolor{red}{1.247}        &\textcolor{red}{1.908} &\textcolor{red}{1.992} &\textcolor{red}{2.264} &3.979 &6.475 &8.348 &\textcolor{red}{8.458} &
            & iTransformer~\cite{nie2024itransformer}   & \textcolor{red}{3.234}       &\textcolor{red}{4.948} &5.745 &21.455 &36.845 &40.347 \\
            \cmidrule(lr){2-18}
            & \fate \textbf{ (Ours) }                             & \textbf{1.183} & \textbf{1.530} & \textbf{1.920} & \textbf{2.041} & \textbf{3.180} & \textbf{5.496} & \textbf{6.677} & \textbf{8.185} &
            & \fate \textbf{ (Ours) }                             & \textbf{3.196} & \textbf{4.335} & \textbf{4.925} & \textbf{19.927} & \textbf{32.454} & \textbf{36.309} \\
           
            \bottomrule
        \end{tabular}
    \end{adjustbox}
    \vspace{1em}
    \label{tab4}
\end{table*}

%% file: Tables/combined_table.tex
\begin{table}[htbp]
\centering
\small
\vspace{1em}
\caption{Comparison of \mae and \mse on temperature prediction across diverse real-world multivariate time-series datasets. The best performing results are highlighted in \textbf{bold} and the second best are marked in \textcolor{red}{red} for clarity.}
\setlength{\tabcolsep}{6pt}
\begin{adjustbox}{max width=\linewidth}
\begin{tabular}{l|cc|cc|cc|cc}
\toprule
\textbf{Model} & \multicolumn{2}{c|}{ETTH1} & \multicolumn{2}{c|}{Traffic} & \multicolumn{2}{c|}{Weather5K} & \multicolumn{2}{c}{ETTM2} \\
& \textbf{MAE} & \textbf{MSE} & \textbf{MAE} & \textbf{MSE} & \textbf{MAE} & \textbf{MSE} & \textbf{MAE} & \textbf{MSE} \\
\midrule
FATE (Ours)        & \textbf{0.381} & 0.377 & \textcolor{red}{0.254} & \textbf{0.349} & \textbf{0.179} & \textbf{0.128} & \textbf{0.221} & \textbf{0.151} \\
CI-TSMixer~\cite{TSMixer}   & \textcolor{red}{0.398} & \textbf{0.368} & 0.278 & \textcolor{red}{0.356} & \textcolor{red}{0.197} & \textcolor{red}{0.146} & \textcolor{red}{0.255} & \textcolor{red}{0.164} \\
PatchTST~\cite{li2023patchtst}   & 0.400 & \textcolor{red}{0.370} & \textbf{0.249} & 0.360 & 0.198 & 0.149 & 0.256 & 0.166 \\
DLinear~\cite{zeng2023dlinear}   & 0.399 & 0.375 & 0.282 & 0.410 & 0.237 & 0.176 & 0.260 & 0.167 \\
FEDformer~\cite{zhou2022fedformer} & 0.419 & 0.376 & 0.366 & 0.587 & 0.296 & 0.217 & 0.287 & 0.203 \\
Autoformer~\cite{wu2021autoformer} & 0.459 & 0.449 & 0.388 & 0.613 & 0.336 & 0.266 & 0.339 & 0.255 \\
Informer~\cite{zhou2021informer}  & 0.713 & 0.865 & 0.391 & 0.719 & 0.384 & 0.300 & 0.453 & 0.365 \\
\bottomrule
\end{tabular}
\end{adjustbox}
\vspace{-1em}
\label{tab:combined_4}
\end{table}

%% file: ICLR/sec/5_conclusion.tex
\section{Outlook and Future Directions}

The strong empirical performance of \textsc{FATE} opens multiple avenues for advancing spatio-temporal forecasting.

\textbf{Scaling to global and ultra-long horizons.}
While \textsc{FATE} performs strongly on regional datasets (Table~\ref{tab4}), scaling to continental or global domains requires optimized training and inference. Future work may explore hierarchical or distributed focal-modulation architectures to retain interpretability while handling millions of spatial points over decades of data.
\textbf{Richer variables and cross-domain fusion.}
Current experiments emphasize temperature and standard meteorological features (Table~\ref{tab:combined_4}). Adding variables such as precipitation, aerosols, oceanic indices, or soil moisture and fusing satellite imagery, reanalysis products, and socio-economic data could enhance predictive power and policy relevance.
\textbf{Self-supervised pretraining.}
Unlabeled climate data motivates self-supervised learning tailored to the focal-tensor setup. Objectives like contrastive or masked prediction can enrich spatio-temporal representations, improve robustness, and reduce dependence on labeled data.
\textbf{Physics-informed inductive biases.}
Incorporating physical constraints e.g., conservation laws or dynamical couplings into focal-modulation blocks may improve physical plausibility and reduce extrapolation error (Appendix~\S\ref{app:corr}). Hybrid integration with NWP ensembles is a promising future direction.
\textbf{Efficiency and real-time inference.}
Though efficient, \textsc{FATE} remains costlier than linear baselines. Techniques such as tensor compression, sparse kernels, or adaptive focal levels could enable lightweight, real-time variants for edge or on-device use.
\textbf{Decision-support and societal impact.}
Translating forecasts into actionable insights for agriculture, energy, and disaster response remains a key challenge. Interpretable modulation maps (Figure~\ref{fig:NewYork}) and tailored visualizations can foster trust and support decision-making.

\textbf{Summary.}
The tensorized focal-modulation design of \textsc{FATE} offers a scalable, extensible foundation for climate forecasting. Future extensions across scale, modality, physics, and application position it as a comprehensive tool for sustainable development.

\section{Conclusion}
In this study, we introduced the \textit{Focal-Modulated Tensorized Encoder} (\fate), a framework designed to capture complex spatiotemporal dependencies in climate data. By leveraging tensorized focal modulation, \fate effectively models multi-scale interactions across time, space, and climate parameters. We evaluated \fate on seven diverse real-world multivariate time series datasets, consistently achieving state-of-the-art performance. Additionally, we proposed head-wise and city-wise modulation scores to enhance interpretability and conducted ablation studies to quantify their impact. This work provides a foundation for informed climate policy decisions and broader applications that exploit 3D tensor-structured data.

\section*{Limitations}
Our current evaluation focuses on temperature and related climate variables within mid-scale regional datasets. Extending \fate to additional meteorological variables and global-scale grids is a direction for future work. While \fate introduces modest computational overhead (trainable on a single A100 GPU), it remains practical for deployment and can be further optimized for edge or real-time applications. These limitations are operational rather than conceptual.